\documentclass[11pt]{article}
\usepackage{booktabs}

\usepackage[final]{acl}

\usepackage{times}
\usepackage{latexsym}
\usepackage{algorithm}
\usepackage{algpseudocode}
\usepackage{amsmath,amssymb,amsfonts}
\usepackage{multirow}

\usepackage[T1]{fontenc}

\usepackage[utf8]{inputenc}
\usepackage[table,xcdraw]{xcolor}

\usepackage{microtype}

\usepackage{inconsolata}
\usepackage[utf8]{inputenc}
\usepackage{geometry}
\usepackage{tcolorbox}
\usepackage{enumitem}

\usepackage{graphicx}
\usepackage{booktabs}
\usepackage{tabularx}
\usepackage{array}

\title{Beyond Local vs. External: A Game-Theoretic Framework for Trustworthy Knowledge Acquisition}

\author{
 \textbf{Rujing Yao\textsuperscript{1}\textsuperscript{*}},
 \textbf{Yufei Shi\textsuperscript{2}\textsuperscript{*}},
 \textbf{Yang Wu\textsuperscript{3}},
 \textbf{Ang Li\textsuperscript{4}},
 \textbf{Zhuoren Jiang\textsuperscript{4}},\\
 \textbf{Xiaofeng Wang\textsuperscript{5}},
 \textbf{Haixu Tang\textsuperscript{6}},
 \textbf{Xiaozhong Liu\textsuperscript{3}\textsuperscript{$\dagger$}}
\\
 \textsuperscript{1}Nankai University, China 
 \textsuperscript{2}The Hong Kong Polytechnic University, China \\
 \textsuperscript{3}Worcester Polytechnic Institute, USA
 \textsuperscript{4}Zhejiang University, China \\ 
 \textsuperscript{5}Nanyang Technological University, Singapore
 \textsuperscript{6}Indiana University Bloomington, USA
\\
\small\texttt{rjyao@mail.nankai.edu.cn, yufei1999.shi@connect.polyu.hk, ywu19@wpi.edu,}\\
\small\texttt{\{leeyon, jiangzhuoren\}@zju.edu.cn, xiaofeng.wang@ntu.edu.sg, hatang@iu.edu, xliu14@wpi.edu}}

\begin{document}
\maketitle
\begin{abstract}
Cloud-hosted Large Language Models (LLMs) offer unmatched reasoning capabilities and dynamic knowledge, yet submitting raw queries to these external services risks exposing sensitive user intent. Conversely, relying exclusively on trusted local models preserves privacy but often compromises answer quality due to limited parameter scale and knowledge. To resolve this dilemma, we propose Game-theoretic Trustworthy Knowledge Acquisition (GTKA), a framework that formulates the trade-off between knowledge utility and privacy as a strategic game. GTKA consists of three components: (i) a privacy-aware sub-query generator that decomposes sensitive intent into generalized, low-risk fragments; (ii) an adversarial reconstruction attacker that attempts to infer the original query from these fragments, providing adaptive leakage signals; and (iii) a trusted local integrator that synthesizes external responses within a secure boundary. By training the generator and attacker in an alternating adversarial manner, GTKA optimizes the sub-query generation policy to maximize knowledge acquisition accuracy while minimizing the reconstructability of the original sensitive intent. To validate our approach, we construct two sensitive-domain benchmarks in the biomedical and legal fields. Extensive experiments demonstrate that GTKA significantly reduces intent leakage compared to state-of-the-art baselines while maintaining high-fidelity answer quality.
\end{abstract}

\renewcommand{\thefootnote}{\fnsymbol{footnote}}
\footnotetext[1]{These authors contributed equally to this work.}
\footnotetext[2]{Corresponding author.}
\renewcommand{\thefootnote}{\arabic{footnote}}

\section{Introduction}
Large Language Models (LLMs) have become foundational tools for knowledge acquisition~\cite{oruganti2023automating, pondel2024ai},  enabling knowledge-intensive organizations, including research institutions, enterprises, and strategic teams, to navigate vast corpora and synthesize information with unprecedented speed~\cite{wang2025survey, aamer2025automating, lai2024leveraging, lewis2020retrieval}. By sharply reducing the time and cost of accessing domain knowledge, LLMs are now integral to daily research workflows~\cite{spatharioti2025effects}. However, this progress introduces a critical risk: many state-of-the-art LLMs are hosted on external cloud platforms (e.g., GPT-5), and submitting raw queries to these services can inadvertently expose high-value organizational intent~\cite{su2024privacy, carlini2021extracting}. Such exposure can result in premature disclosure, idea appropriation, and competitive disadvantage, since intent-bearing queries may be logged, profiled, or inferred by external providers. For instance, consider a biomedical researcher investigating a novel therapeutic target for a specific cancer. A single query to a cloud-based LLM about highly specific compounds or mechanisms could expose their entire research trajectory. In these settings, intent leakage does not merely threaten individual privacy but can compromise organizational competitiveness and long-term value, long before any public disclosure occurs.

\begin{figure}[t]
\setlength{\belowcaptionskip}{-0.3cm}
    \centering
    \includegraphics[width=1\linewidth]{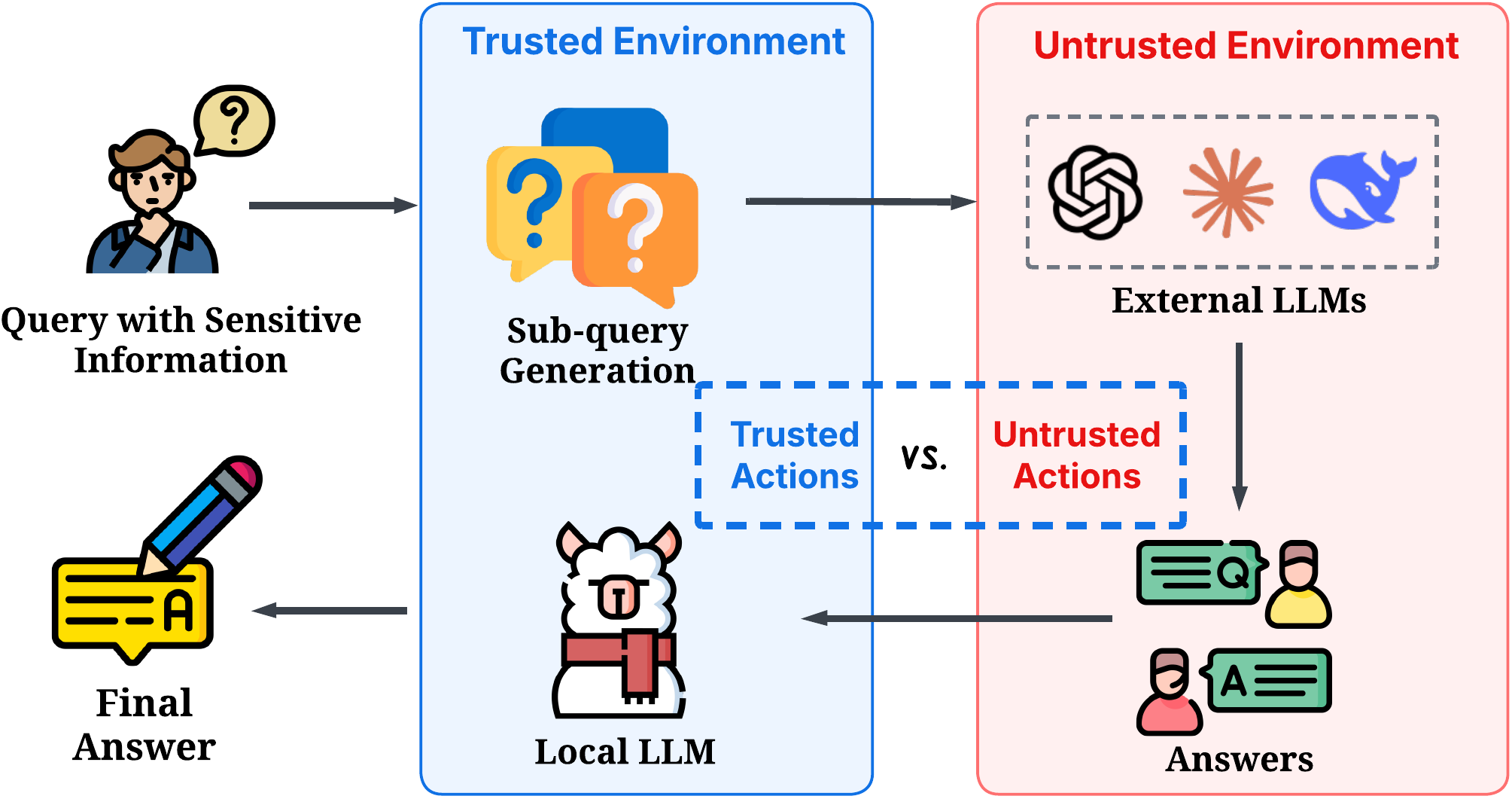}
    \caption{An illustration of our method.}
    \label{fig:fig1}
\end{figure} 

A large and growing literature seeks to curb information leakage from multiple angles~\cite{sweeney2002k}. For example, query-side transformations, such as paraphrasing and semantic obfuscation, attempt to hide intent but create a blunt trade-off: aggressive transformations distort the user's core information need, while subtle changes fail to meaningfully reduce risk~\cite{utpala2023locally, pape2025prompt}. Training-time protections, such as differential privacy, harden model parameters but leave the live query channel exposed~\cite{abadi2016deep, pang2024reconstruction}.

Cryptographic and hardware-based solutions, including Multi-Party Computation (MPC), Homomorphic Encryption (HE), and Trusted Execution Environments (TEEs), offer significantly stronger confidentiality guarantees~\cite{beimel2011secret, chattopadhyay2024secret, pillai2024enhancing, gentry2009fully, ohrimenko2016oblivious}. However, unlike client-side semantic strategies, these methods require fundamental changes to the service hosting the model. Furthermore, MPC and HE suffer from prohibitive computational overheads. While TEEs provide a promising and efficient hardware-backed alternative, they necessitate migrating current deployments into encrypted virtual machines, representing a widespread infrastructure overhaul likely requiring years to mature~\cite{mo2021ppfl}. Moreover, even if the primary LLM is secured, downstream external tools may still fall outside the trusted perimeter.

To address these challenges, we propose a novel semantics-level framework that simultaneously leverages cloud-hosted LLMs for knowledge acquisition while safeguarding research intent from disclosure. As illustrated in Figure~\ref{fig:fig1}, given a sensitive user query, we first strategically decompose it into a set of semantically fragmented, low-risk subqueries. These subqueries are then dispatched to multiple, diverse cloud-hosted LLM endpoints to acquire the necessary external knowledge. Finally, a local LLM, which operates within the user's trusted environment and is the only component to see the original sensitive context, aggregates the knowledge fragments returned from the external LLM services. A coherent, high-quality final answer is obtained. Compared with prior approaches, our method offers two key advantages. First, it leverages existing hosted LLMs without modification, thereby avoiding the complexity and inefficiency of cryptographic or infrastructure-heavy solutions. Second, it fundamentally enhances user trust and security by ensuring that raw queries containing sensitive intent are never directly exposed to any external, untrusted service.

The contributions are summarized as follows:
\begin{itemize}
    \item We propose a game-theoretic framework for trustworthy knowledge acquisition (GTKA), which enables high-quality knowledge acquisition from cloud-hosted LLMs while safeguarding sensitive user intent.
    
    \item We cast the privacy--utility trade-off as an adversarial game between a privacy-aware sub-query generator and an adversarial reconstruction attacker. We optimize the sub-query policy via alternating adversarial training to jointly maximize answer quality and minimize intent reconstructability.
    
    \item We construct two sensitive-domain benchmarks focused on biomedical and legal contexts. Extensive evaluations demonstrate that GTKA substantially reduces intent leakage against strong baselines while maintaining high-fidelity answer quality.
\end{itemize}

\section{Related Work}

\subsection{Privacy Protection for Large Language Models}
As large language models (LLMs) increasingly mediate human–AI knowledge exchange, concerns over privacy leakage have become central to their responsible deployment~\cite{das2025security,chen2025survey}. Prior research has explored diverse mechanisms to mitigate information exposure across the entire LLM lifecycle.  
Early work emphasizes parameter-level protections such as differential privacy and secure aggregation to prevent memorization of sensitive data during training~\cite{abadi2016deep,lyu2020threats}. Subsequent studies extend protection to inference and deployment, employing confidential execution environments and encrypted federated pipelines to safeguard user data from server-side inspection~\cite{yin2021comprehensive,zhao2024privacy}.  
More recently, the focus has shifted toward the interaction layer, where user prompts themselves become a major source of leakage risk. Semantic rewriting and paraphrastic obfuscation~\cite{utpala2023locally}, along with selective local sanitization~\cite{kan2023protecting}, attempt to conceal intent before transmission to cloud-hosted models. While effective in isolation, these strategies remain either static, failing to adapt to the evolving inference behavior of external models, or costly, due to cryptographic or architectural constraints.

\subsection{Game-Theoretic Formulations}
Game theory provides a principled foundation for analyzing strategic interactions in machine learning and multi-agent systems. Classical formulations such as zero-sum and Nash games underpin adversarial training and robust optimization, where competing agents iteratively minimize and maximize shared objectives~\cite{goodfellow2014explaining,madry2017towards}. Beyond these symmetric settings, hierarchical formulations, most notably Stackelberg games~\cite{bacsar1998dynamic,de2016machine}, model leader–follower dynamics and have become central to mechanism design, security, and privacy-aware learning. Recent studies have extended Stackelberg formulations to modern AI and large language models (LLMs). Theoretically, differentiable Stackelberg solvers now enable gradient-based optimization through implicit equilibria~\cite{fiez2020implicit,li2020end}. In alignment, STA-RLHF~\cite{makar2024sta} interprets reinforcement learning from human feedback as a leader–follower interaction between the policy and reward model, while SGPO~\cite{chu2025stackelberg} leverages Stackelberg equilibrium for more data-efficient preference optimization. Beyond alignment, Stackelberg frameworks have been adopted for LLM detoxification~\cite{xie2024learning}, attacker–defender modeling in jailbreak prevention~\cite{han2025dynamic}, and robust federated learning under adversarial or mixed attacks~\cite{li2024meta}.

\section{Methodology}
In this section, we present GTKA, a framework designed to balance quality and privacy in local-external LLM collaborative inference. We begin by formally defining the problem setting. Subsequently, we elaborate on the proposed method.
\subsection{Problem Formulation}

\paragraph{Preliminaries.}
We consider a setting involving two distinct agents: a \textit{trusted local LLM} ($\mathcal{M}_{loc}$) and an \textit{untrusted external LLM} ($\mathcal{M}_{ext}$). 
$\mathcal{M}_{loc}$ operates within a secure boundary but is constrained by limited model parameters and knowledge. 
Conversely, $\mathcal{M}_{ext}$ (e.g., GPT-5) possesses superior reasoning capabilities and dynamic knowledge but poses potential privacy risks as it resides in an external, untrusted environment.

\paragraph{Task Definition.}

Given a user's sensitive query $q$, the primary objective is to acquire a high-quality answer $\hat{a}$ by leveraging $\mathcal{M}_{ext}$, without exposing the privacy of $q$. 
Directly querying $\mathcal{M}_{ext}$ with $q$ maximizes quality but compromises privacy, whereas relying solely on $\mathcal{M}_{loc}$ ensures privacy but often yields suboptimal responses. 
Formally, we employ a trusted local LLM to generate a sequence of $n$ low-leakage sub-queries $\mathcal{S} = \{s_1, s_2, \dots, s_n\}$. 
These sub-queries are sent to $\mathcal{M}_{ext}$ to obtain a corresponding set of responses $\mathcal{A} = \{a_1, a_2, \dots, a_n\}$. Finally, a trusted local integrator aggregates the sub-queries, external responses, and original query $(\mathcal{S}, \mathcal{A}, q)$ to synthesize the final answer $\hat{a}$.

\subsection{The GTKA Framework}
\begin{figure*}[htbp]
\setlength{\belowcaptionskip}{-0cm}
    \centering
    \includegraphics[width=1\linewidth]{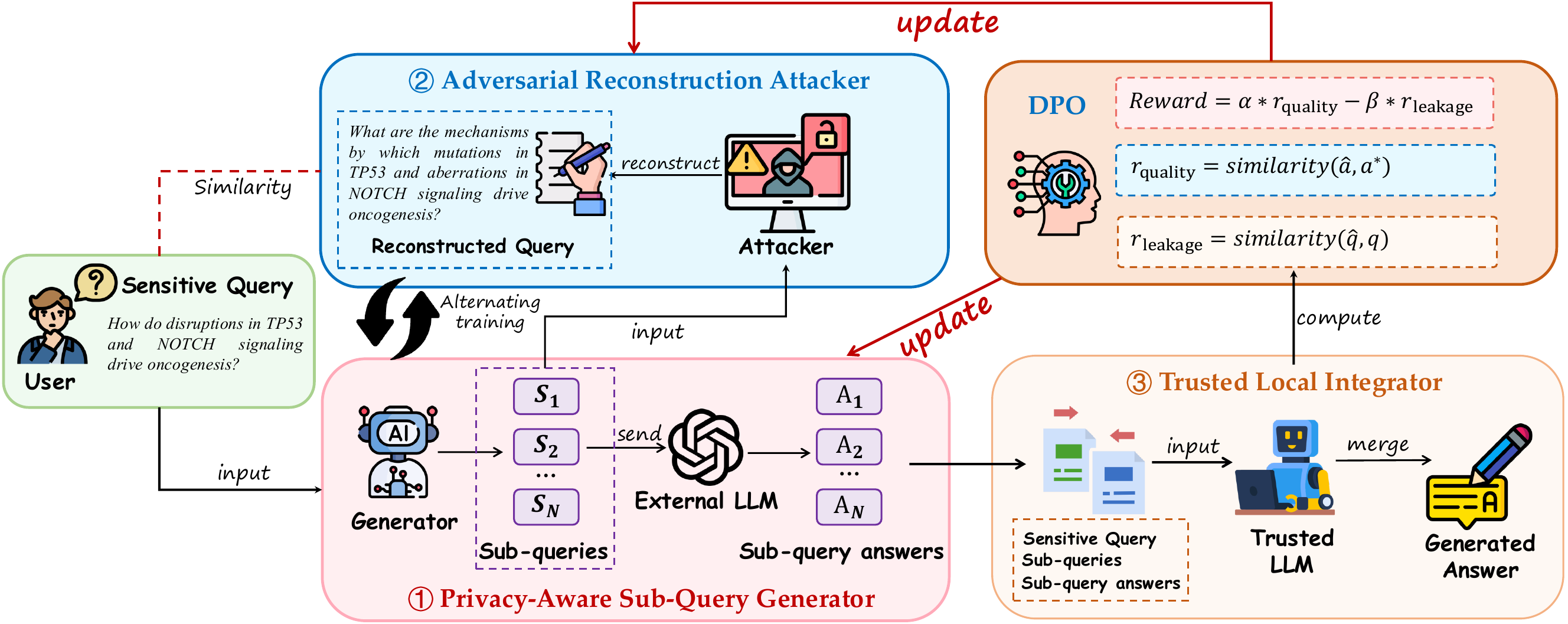}
    \caption{The overall framework of GTKA. The network consists of three modules: Privacy-Aware Sub-Query Generator, Adversarial Reconstruction Attacker, and Trusted Local Integrator.}
    \label{fig:fig2}
\end{figure*}

As illustrated in Figure~\ref{fig:fig2}, we propose a Game-theoretic Trustworthy Knowledge Acquisition (GTKA) framework. The framework consists of three modules: Privacy-Aware Sub-Query Generator, Adversarial Reconstruction Attacker, and Trusted Local Integrator.

\subsubsection{Privacy-Aware Sub-Query Generator}
The privacy-aware sub-query generator aims to produce a set of low-leakage sub-queries that are maximally useful for downstream knowledge acquisition while minimizing exposure of the original sensitive intent. Let the privacy-aware sub-query generator be denoted as $G_\theta$, which is deployed within the trusted local environment. 

Rather than paraphrasing the sensitive query $q$, $G_\theta$ generates a set of generalized, low-leakage sub-queries that preserve what information is needed to answer the request, such as key concepts to clarify, criteria to apply, constraints that do not reveal identities, and commonly used reasoning patterns, while masking sensitive entities and context-specific identifiers.

Specifically, for each user input $q$, the generator $G_\theta$ samples $K$ candidate sub-query groups $\{\mathcal{S}^{(1)},\ldots,\mathcal{S}^{(K)}\}$, where each group $\mathcal{S}^{(k)}=\{s^{(k)}_1,\ldots,s^{(k)}_n\}$ contains $n$ sub-queries. For each candidate group $\mathcal{S}^{(k)}$, we send its sub-queries to the external LLM and obtain responses $\mathcal{A}^{(k)}=\{a^{(k)}_1,\ldots,a^{(k)}_n\}$, where $a^{(k)}_i=\mathcal{M}_{ext}(s^{(k)}_i)$. We then evaluate each group using a quality--privacy reward defined as
\begin{equation}
\begin{aligned}
R(q,\mathcal{S}^{(k)}) &=
\alpha\, \text{Quality}(\hat{a}^{(k)},a^\star) \\
&\quad - \beta\, \text{Leakage}(q,\mathcal{S}^{(k)}).
\end{aligned}
\end{equation}
where $\hat{a}^{(k)}$ denotes the locally integrated answer produced from $(q,\mathcal{S}^{(k)},\mathcal{A}^{(k)})$, $a^\star$ is the reference answer, and $\alpha,\beta$ control the quality--privacy trade-off. 

The Quality term is defined as semantic similarity between the locally integrated answer
$\hat{a}^{(k)}$ and the reference answer $a^\star$:
\begin{equation}
\text{Quality}(\hat{a}^{(k)}, a^\star)=\text{Sim}(\hat{a}^{(k)}, a^\star).
\end{equation}

The leakage term $\text{Leakage}(q,\mathcal{S}^{(k)})$ measures the privacy risk induced by releasing the candidate sub-query group $\mathcal{S}^{(k)}$, as estimated by the adversarial reconstruction attacker described in Section~\ref{attacker}.

Based on this reward, we convert the $K$ sampled candidate groups into pairwise preferences and train $G_\theta$
with Direct Preference Optimization (DPO)~\cite{rafailov2023direct}.

\paragraph{Preference Construction.}
For each query $q$, we compute rewards $\{R(q,\mathcal{S}^{(k)})\}_{k=1}^{K}$ for all sampled candidate groups
and construct a preference pair by selecting a higher-reward group $\mathcal{S}^+$ and a lower-reward group
$\mathcal{S}^-$. Concretely,
\begin{equation}
\mathcal{S}^{+} = \arg\max_{k}\; R\!\left(q,\mathcal{S}^{(k)}\right).
\end{equation}

\begin{equation}
\mathcal{S}^{-} = \arg\min_{k}\; R\!\left(q,\mathcal{S}^{(k)}\right).
\end{equation}

\paragraph{DPO Optimization.}
Let $\pi_\theta(\mathcal{S}\!\mid\! q)$ denote the generator policy, and let $\pi_{\text{ref}}(\mathcal{S}\!\mid\! q)$
be a fixed reference policy initialized from the pre-trained generator.
Given each preference pair $(q,\mathcal{S}^+,\mathcal{S}^-)$, DPO updates $G_\theta$ to assign higher likelihood
to the preferred group $\mathcal{S}^+$ than the less-preferred group $\mathcal{S}^-$.

\begin{algorithm}[t]
\caption{GTKA Training}
\label{alg:alg1}
\begin{algorithmic}[1]
\Require Dataset $\mathcal{D}$, Generator $G_\theta$, Attacker $A_\phi$, Integrator $\mathcal{I}$, External LLM $\mathcal{M}_{ext}$.
\Ensure Optimized $G_{\theta}$ and $A_{\phi}$.
\State Initialize $\theta, \phi$; $\pi_{\text{ref}} \leftarrow G_\theta$.
\For{iteration $t = 1, \dots, T$}
    \State Sample a batch of queries $\mathcal{B}_q \subset \mathcal{D}$.
    
    \State \textcolor{gray}{// Generate $K$ candidates for each query in batch}
    \State $\mathcal{B}_{\mathcal{S}} \gets \{ (q, \{\mathcal{S}^{(k)}\}_{k=1}^K) \mid q \in \mathcal{B}_q, \mathcal{S}^{(k)} \sim G_\theta(q) \}$.
    
    \State \textcolor{gray}{// Optimize Attacker}
    \State Update $\phi \gets \operatorname{SFT}(A_\phi, \mathcal{B}_{\mathcal{S}})$ to maximize reconstruction likelihood.
    
    \State \textcolor{gray}{// Optimize Generator}
    \State $\mathcal{D}_{\text{pref}} \gets \emptyset$
    \For{each $(q, \{\mathcal{S}^{(k)}\}_{k=1}^K) \in \mathcal{B}_{\mathcal{S}}$}
        \State Compute rewards $R^{(k)}$ for each candidate using \textbf{updated} $A_\phi$.
        \State Construct pair $(\mathcal{S}^+, \mathcal{S}^-)$; add to $\mathcal{D}_{\text{pref}}$.
    \EndFor
    \State Update $\theta \gets \operatorname{DPO}(G_\theta, \pi_{\text{ref}}, \mathcal{D}_{\text{pref}})$.
\EndFor
\State \Return $G_\theta, A_\phi$
\end{algorithmic}
\end{algorithm}

\subsubsection{Adversarial Reconstruction Attacker}
\label{attacker}

To simulate realistic privacy threats, we introduce an Adversarial Reconstruction Attacker, denoted as $A_\phi$. This module mimics an adversary who intercepts the sub-queries released to an untrusted external service and attempts to infer the user's private intent. We formulate this attack as a reconstruction problem, where the attacker observes the candidate sub-query group $\mathcal{S}^{(k)}$ and generates a reconstructed query $\hat{q}^{(k)}$:
\begin{equation}
\hat{q}^{(k)} = A_\phi(\mathcal{S}^{(k)}).
\end{equation}

We then quantify privacy leakage by measuring the semantic similarity between the reconstructed query $\hat{q}^{(k)}$ and the ground-truth query $q$:
\begin{equation}
\text{Leakage}(q, \mathcal{S}^{(k)}) 
= \text{Sim}(q, \hat{q}^{(k)}).
\end{equation}

A higher leakage score indicates a more severe breach of privacy.

In the paper, the attacker $A_\phi$ is implemented as an LLM and optimized via supervised fine-tuning. During GTKA training, the attacker provides an adaptive leakage signal that penalizes candidate sub-query groups from which the original query can be easily reconstructed. This adversarial mechanism encourages the generator to produce sub-queries that reveal less information and are
more difficult to reverse-engineer into the original query. We train the generator and the attacker in an alternating manner. The detailed training procedure is presented in Algorithm~\ref{alg:alg1}.

\subsubsection{Trusted Local Integrator}
After the alternating training of the generator $G_\theta$ and the attacker $A_\phi$, we fix the trained
$G_\theta$ and use it at inference time to generate a sub-query group $\mathcal{S}=G_\theta(q)$ for each sensitive user
query $q$. These sub-queries are then dispatched to the external LLM to obtain a set of responses
$\mathcal{A}=\{a_1,\ldots,a_n\}$, while the original query $q$ remains strictly within the trusted local environment.

After obtaining external responses, GTKA performs answer synthesis locally to ensure that the sensitive query $q$ never leaves the trusted boundary. The trusted local integrator $\mathcal{I}$ aggregates the original query, sub-queries, and external responses to produce the final answer:
\begin{equation}
\hat{a}=\mathcal{I}(q,\mathcal{S},\mathcal{A}).
\end{equation}

In our framework, $\mathcal{I}$ is implemented as a trusted local LLM $\mathcal{M}_{loc}$ that runs entirely within the secure boundary. It treats $\mathcal{A}$ as external knowledge and synthesizes a final answer that directly addresses $q$. Only the sub-queries $\mathcal{S}$ are sent to the untrusted external model, while the final reasoning that grounds external information in the user's original intent is carried out locally. As a result, GTKA provides privacy by construction because $q$ is never disclosed to $\mathcal{M}_{ext}$, yet it can still leverage $\mathcal{M}_{ext}$ for broad and up-to-date knowledge.

\section{Experiments}
\subsection{Dataset}

To comprehensively evaluate the effectiveness and generalizability of our proposed method, we conduct experiments on two sensitive-domain QA datasets from the biomedical and legal domains, as both involve sensitive intents and high-stakes decision making where privacy leakage can cause real harm. Existing biomedical and legal QA datasets are often difficult to trace back to their source documents, which prevents rigorous evaluation of privacy leakage. Therefore, we construct two domain-specific QA datasets in this work, BioQA and LawQA.

BioQA is a high-quality question answering (QA) dataset in the biomedical domain, constructed from PubMed articles. For both datasets, we use an 8:2 split for training and testing. Dataset statistics are reported in Table~\ref{tab:dataset}, and the dataset construction process is detailed in Appendix~\ref{app:dataset_construction}.

\begin{table}[htbp]
\centering
\small
\resizebox{\columnwidth}{!}{
\begin{tabular}{ccc}
\toprule
\textbf{Type} & \textbf{BioQA} & \textbf{LawQA} \\
\midrule
Total                & 12,876 & 12,575 \\
Train                & 10,301 & 10,060 \\
Test                 & 2,575  & 2,515  \\
Avg. Q length (words)& 13.90    & 128.18     \\
Avg. A length (words)& 52.42     & 146.95     \\
\bottomrule
\end{tabular}
}
\caption{Statistics of the two datasets. ``Avg. Q length'' and ``Avg. A length'' denote the average number of words per question and answer, respectively.}
\label{tab:dataset}
\end{table}

\begin{table*}[htbp]
\centering
\small
\definecolor{MyGreen}{RGB}{34, 139, 34}
\setlength{\tabcolsep}{5pt}
\renewcommand{\arraystretch}{1.12}
\begin{tabularx}{\textwidth}{@{}>{\centering\arraybackslash}m{0.28\textwidth}*{5}{>{\centering\arraybackslash}X}@{\hspace{14pt}}*{3}{>{\centering\arraybackslash}X}@{}}
\toprule
\multirow{2}{*}[-0.3ex]{\centering\arraybackslash\textbf{Method}} &
\multicolumn{5}{c}{\textbf{Knowledge Acquisition ($\uparrow$)}} &
\multicolumn{3}{c}{\textbf{Intent Leakage}($\downarrow$)} \\
\cmidrule(lr){2-6}\cmidrule(lr){7-9}
& R-1 & R-2 & R-L & METEOR & BERTScore
& ASR@1 & ASR@3 & MRR \\
\midrule
Local-Only (Qwen2.5-3B-Instruct) & 16.12 & 3.57 &12.61 & 20.40 & 80.27 & / & / & / \\
External-Only (Qwen-Turbo) & 24.96 & 6.50 & 19.17 & 25.21 & 83.99 & / & / & / \\
PP-TS~\citep{kan2023protecting} & 16.58 & 3.63 & 12.78 & 20.56 & 80.52 & 21.48 & 39.65 & 31.60 \\
IOI~\citep{yao2024privacy} & 17.86 & 3.98 & 14.36 & 21.24 & 81.15 & 19.89 & 37.99 & 30.03 \\
Hard-PO~\citep{pape2025prompt} & 16.74 & 3.71 & 13.72 & 21.85 & 81.58 & 19.33 & 37.02 & 29.21 \\
Soft-PO~\citep{pape2025prompt} & 19.42 & 4.56 & 14.85 & 22.18 & 81.92 & 20.56 & 38.72 & 30.77 \\
\midrule
\textbf{GTKA} & \textbf{21.35} & \textbf{5.18} & \textbf{16.42} & \textbf{23.46} & \textbf{82.85} & \textbf{18.41}  & \textbf{36.14} & \textbf{28.43} \\
\bottomrule
\end{tabularx}
\caption{Performance comparison on the BioQA dataset with Qwen2.5-3B-Instruct (trusted local model) and Qwen-Turbo (untrusted external model). $\uparrow$ indicates higher scores are better, while $\downarrow$ indicates lower scores are better.}
\label{tab:results_BioQA_3B}
\end{table*}

\begin{table*}[htbp]
\centering
\small
\definecolor{MyGreen}{RGB}{34, 139, 34}
\setlength{\tabcolsep}{5pt}
\renewcommand{\arraystretch}{1.12}
\begin{tabularx}{\textwidth}{@{}>{\centering\arraybackslash}m{0.24\textwidth}*{8}{>{\centering\arraybackslash}X}@{}}
\toprule
\multirow{2}{*}[-0.3ex]{\centering\arraybackslash\textbf{Method}} &
\multicolumn{5}{c}{\textbf{Knowledge Acquisition ($\uparrow$)}} &
\multicolumn{3}{c}{\textbf{Intent Leakage}($\downarrow$)} \\
\cmidrule(lr){2-6}\cmidrule(lr){7-9}
& R-1 & R-2 & R-L & METEOR & BERTScore
& ASR@1 & ASR@3 & MRR \\
\midrule
Local-Only (Llama-3.1-8B) & 16.35 & 3.26 &11.62 & 18.17 & 81.02 & / & / & / \\
External-Only (GPT‑4o mini) & 21.41 & 5.38 & 16.34 & 23.83 & 84.10 & / & / & /\\
PP-TS~\citep{kan2023protecting} & 15.62 &3.25 & 11.76 & 19.24 & 81.67 & 14.16 & 31.98 & 25.76 \\
IOI~\citep{yao2024privacy} & 16.28 & 3.54 & 12.24 & 19.72 & 82.86 & 12.71 & 30.14 & 24.16 \\
Hard-PO~\citep{pape2025prompt} & 17.04 & 3.98 & 13.47 & 20.61 & 82.94 & 11.99 & 29.37 & 23.26 \\
Soft-PO~\citep{pape2025prompt} & 17.25 & 4.05 & 13.62 & 20.78 & 83.34 & 13.44 & 31.05 & 24.78 \\
\midrule
\textbf{GTKA} & \textbf{18.79} & \textbf{4.52} & \textbf{14.38} & \textbf{21.94} & \textbf{83.72} & \textbf{11.12} & \textbf{28.57} & \textbf{22.52} \\
\bottomrule
\end{tabularx}
\caption{Performance comparison on the BioQA dataset with Llama-3.1-8B (trusted local model) and GPT‑4o mini (untrusted external model). $\uparrow$ indicates higher scores are better, while $\downarrow$ indicates lower scores are better.}
\label{tab:results_BioQA_8B}
\end{table*}

\begin{table*}[t]
\centering
\small
\definecolor{MyGreen}{RGB}{34, 139, 34}
\setlength{\tabcolsep}{5pt}
\renewcommand{\arraystretch}{1.12}
\begin{tabularx}{\textwidth}{@{}>{\centering\arraybackslash}m{0.28\textwidth}*{5}{>{\centering\arraybackslash}X}@{\hspace{14pt}}*{3}{>{\centering\arraybackslash}X}@{}}
\toprule
\multirow{2}{*}[-0.3ex]{\centering\arraybackslash\textbf{Method}} &
\multicolumn{5}{c}{\textbf{Knowledge Acquisition ($\uparrow$)}} &
\multicolumn{3}{c}{\textbf{Intent Leakage}($\downarrow$)} \\
\cmidrule(lr){2-6}\cmidrule(lr){7-9}
& R-1 & R-2 & R-L & METEOR & BERTScore
& ASR@1 & ASR@3 & MRR \\
\midrule
Local-Only (Qwen2.5-3B-Instruct) & 39.99 & 17.90 &28.74 & 43.94 & 81.57 & / & / & / \\
External-Only (Qwen-Turbo) & 45.84 & 21.06 & 32.49 & 45.33 & 85.80 & / & / & / \\
PP-TS~\citep{kan2023protecting} & 40.15 & 17.92 & 28.77 & 43.72 & 81.96 & 23.41 & 46.19 & 36.05 \\
IOI~\citep{yao2024privacy} & 40.72 & 18.08 & 29.02 & 43.83 & 82.41 & 21.94 & 44.34 & 34.41 \\
Hard-PO~\citep{pape2025prompt} & 41.24 & 18.25 & 29.18 & 44.03 & 82.35 & 21.24 & 43.49 & 33.70 \\
Soft-PO~\citep{pape2025prompt} & 42.56 & 18.42 & 29.35 & 44.56 & 83.87 & 22.57 & 45.22 & 35.07 \\
\midrule
\textbf{GTKA} & \textbf{43.52} & \textbf{19.86} & \textbf{30.84} & \textbf{44.91} & \textbf{84.69} & \textbf{20.35} & \textbf{42.56} & \textbf{32.74} \\
\bottomrule
\end{tabularx}
\caption{Performance comparison on the LawQA dataset with Qwen2.5-3B-Instruct (trusted local model) and Qwen-Turbo (untrusted external model). $\uparrow$ indicates higher scores are better. $\downarrow$ indicates lower scores are better.}
\label{tab:results_LawQA_3B}
\end{table*}

\begin{table*}[t]
\centering
\small
\definecolor{MyGreen}{RGB}{34, 139, 34}
\setlength{\tabcolsep}{5pt}
\renewcommand{\arraystretch}{1.12}
\begin{tabularx}{\textwidth}{@{}>{\centering\arraybackslash}m{0.24\textwidth}*{8}{>{\centering\arraybackslash}X}@{}}
\toprule
\multirow{2}{*}[-0.3ex]{\centering\arraybackslash\textbf{Method}} &
\multicolumn{5}{c}{\textbf{Knowledge Acquisition ($\uparrow$)}} &
\multicolumn{3}{c}{\textbf{Intent Leakage}($\downarrow$)} \\
\cmidrule(lr){2-6}\cmidrule(lr){7-9}
& R-1 & R-2 & R-L & METEOR & BERTScore
& ASR@1 & ASR@3 & MRR \\
\midrule
Local-Only (Llama-3.1-8B) & 40.49 & 18.64 &28.78 & 43.97 & 81.81 & / & / & /\\
External-Only (GPT‑4o mini) & 44.71 & 20.72 & 30.90 & 46.28 & 85.98 & / & / & / \\
PP-TS~\citep{kan2023protecting} & 38.56 & 16.82 & 26.54 & 42.28 & 82.13 & 25.43 & 48.46 & 37.92 \\
IOI~\citep{yao2024privacy} & 39.72 & 17.56 & 27.48 & 43.12 & 83.08 & 23.98 & 46.71 & 36.27 \\
Hard-PO~\citep{pape2025prompt} & 40.35 & 17.86 & 27.92 & 43.45 & 83.72 & 23.19 & 45.85 & 35.61 \\
Soft-PO~\citep{pape2025prompt} & 40.86 & 18.12 & 28.24 & 43.68 & 84.76 & 24.68 & 47.64 & 37.09 \\
\midrule
\textbf{GTKA} & \textbf{42.67} & \textbf{19.58} & \textbf{29.72} & \textbf{45.36} & \textbf{85.39} & \textbf{22.55} & \textbf{44.93} & \textbf{34.68} \\
\bottomrule
\end{tabularx}
\caption{Performance comparison on the LawQA dataset with Llama-3.1-8B (trusted local model) and GPT‑4o mini (untrusted external model). $\uparrow$ indicates higher scores are better, while $\downarrow$ indicates lower scores are better.}
\label{tab:results_LawQA_8B}
\end{table*}

\subsection{Baselines and Evaluation Metrics}

\textbf{Baselines.} \textbf{Local-Only (Trusted):} Answering the raw query solely with a local, trusted LLM (usually smaller) operating within the secure environment, without calling any external cloud LLMs. \textbf{External-Only (Untrusted):} Directly submitting the raw, unmodified sensitive query to the external cloud LLM. \textbf{PP-TS~\citep{kan2023protecting}:} The framework protects user privacy by locally filtering sensitive information from user inputs before transmitting them to a remote LLM, and subsequently restoring the filtered content in the response. \textbf{IOI~\citep{yao2024privacy}:} The true input instance is combined with a dummy instance to form an obfuscated instance, which is subsequently transmitted to the external LLM. \textbf{Hard-Prompt Obfuscation~\citep{pape2025prompt}:} The method operates in the discrete token space, iteratively replacing tokens to generate a textually distinct yet functionally equivalent hard obfuscated prompt. \textbf{Soft-Prompt Obfuscation~\citep{pape2025prompt}:} The method operates in the continuous embedding space, directly optimizing the prompt's embedding vectors to generate a functionally equivalent soft obfuscated prompt.

\begin{figure*}[htbp]
\setlength{\belowcaptionskip}{-0cm}
    \centering
    \includegraphics[width=0.93\linewidth]{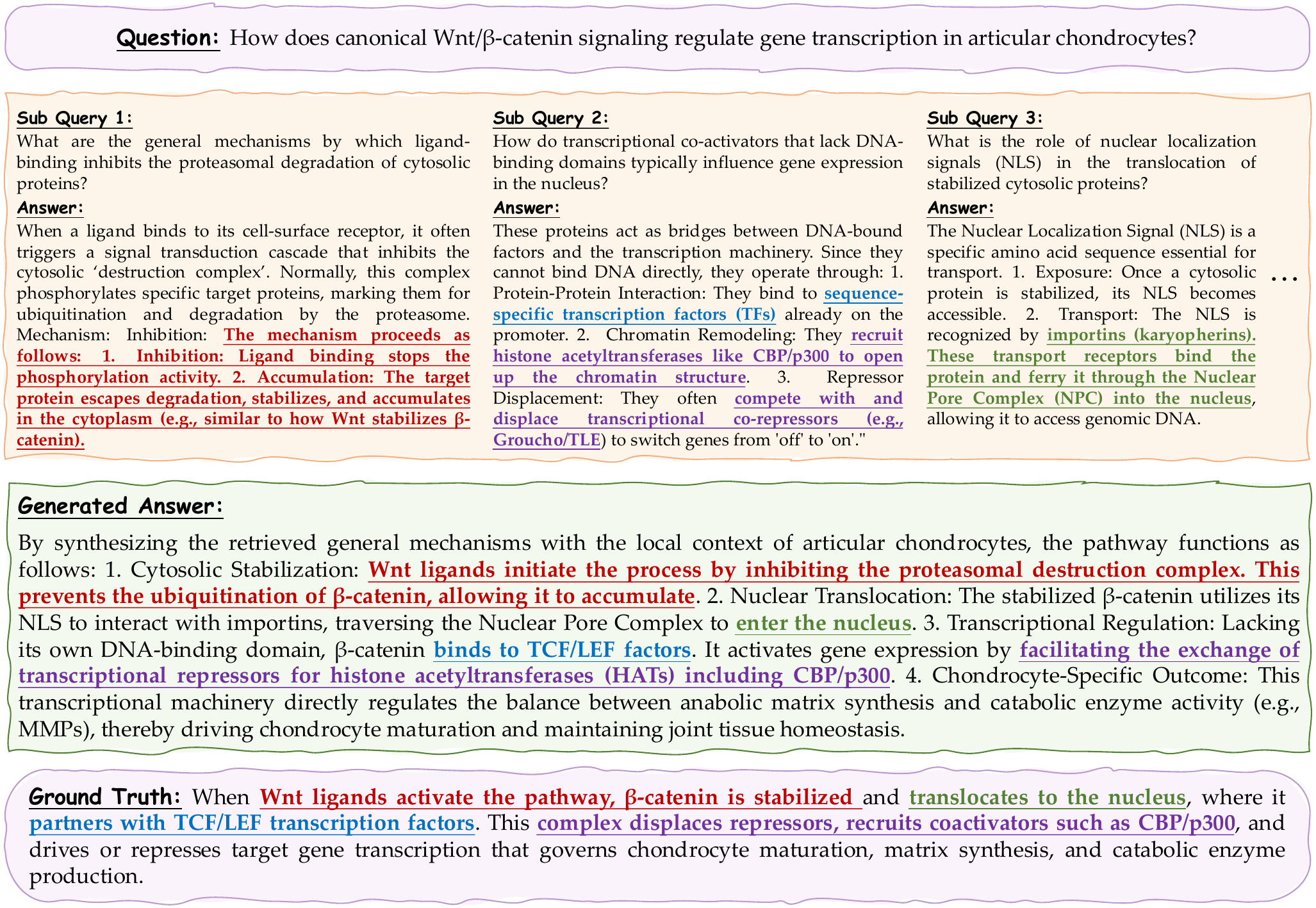}
    \caption{Biomedical case (Text highlighted in the same color denotes semantically corresponding information). Starting from the original query, our proposed GTKA method generates multiple low-leakage sub-queries that avoid revealing the original intent, submits them to an external LLM to obtain intermediate responses, and then uses a trusted local model to securely integrate these responses into the final answer.}
    \label{fig:case_study}
\end{figure*}

\textbf{Evaluation Metrics.} To evaluate our framework comprehensively, we assess all methods along two dimensions: knowledge acquisition and intent leakage, using both automatic and human evaluation.
For Knowledge acquisition, the automatic evaluation reports commonly used QA metrics, including ROUGE-1 (R-1), ROUGE-2 (R-2), ROUGE-L (R-L), METEOR, and BERTScore.

For intent leakage, we adopt two attacker-based automatic metrics: Attack Success Rate (ASR) and Mean Reciprocal Rank (MRR).
Specifically, for each privacy-preserved input, given a predefined candidate set consisting of the true source segment and several decoy segments, the attacker produces a ranked list of candidates according to how well each candidate matches the observed input.
ASR@k is defined as Top-$k$ accuracy, i.e., the fraction of instances where the true segment appears in the top $k$ positions. A higher ASR@k indicates stronger leakage. MRR measures the average reciprocal rank of the true segment in the attacker’s ranking. The detailed calculations are provided in Appendix~\ref{app:metrics}.

For human evaluation, we recruited three domain experts, and each instance was annotated independently by all three experts. For knowledge acquisition, the experts were provided with the question, the model-generated answer, and the gold reference answer, and then rated the response on a 0–5 scale along three dimensions: accuracy, completeness, and adoptability. Specifically, accuracy measures factual correctness against the gold reference, completeness measures coverage of essential points needed to fully address the question, and adoptability measures as-is usability without additional rewriting.

For intent leakage, we conducted a human attack simulation: the experts were given the privacy-preserving query together with a candidate pool consisting of the true source segment and multiple decoy segments, and were asked to rank the candidate segments according to how likely each one was to be the true source based solely on the obfuscated query. We report the average scores across the three experts as the final human evaluation results.

\subsection{Experimental Settings}
To comprehensively evaluate the performance of our proposed GTKA framework, we employ two distinct configurations of base Large Language Models (LLMs). First, we utilize Qwen2.5-3B-Instruct as the local trusted model paired with Qwen-Turbo as the external untrusted model, where the former serves as the backbone for both the generator and the attacker. Second, we adopt Llama-3.1-8B as the local trusted model and GPT-4o-mini as the external model, with Llama-3.1-8B similarly functioning as both the generator and attacker. Additional experimental settings are detailed in Appendix~\ref{app:exp_setting}.

\subsection{Experimental Results}
Tables~\ref{tab:results_BioQA_3B} and~\ref{tab:results_BioQA_8B} report the results on the BioQA dataset under two different local--external model settings. Overall, GTKA maintains strong knowledge acquisition while achieving the lowest intent leakage. The performance on the LawQA dataset is detailed in Tables~\ref{tab:results_LawQA_3B} and~\ref{tab:results_LawQA_8B}. The experimental results demonstrate that GTKA effectively masks sensitive legal intent while successfully obtaining pertinent content from external LLMs, thereby further validating the generalization capability of our framework in the legal domain.

To assess the practical usefulness and privacy protection of our method, we conducted a human evaluation, with results reported in Table~\ref{tab:human_bioqa}. GTKA achieves the strongest performance across the accuracy, completeness, and adoptability metrics. For intent leakage, GTKA also exhibits substantially lower leakage than the compared method. These findings indicate that the sub-queries generated by GTKA effectively mask the original intent from adversaries while still obtaining relevant knowledge from external LLMs and producing high-quality answers.

\begin{table}[htbp]
\centering
\small
\captionsetup{skip=8pt}
\setlength{\tabcolsep}{6pt}
\begin{tabular}{lccc}
\toprule
\multicolumn{4}{c}{\textbf{Human: Knowledge Acquisition ($\uparrow$)}} \\
\cmidrule(lr){1-4}
\textbf{Method} & Accuracy & Completeness & Adoptability \\
\midrule
Soft-PO & 4.17 & 4.03 & 3.96 \\
GTKA   & 4.65 & 4.54 & 4.72 \\
\midrule
\multicolumn{4}{c}{\textbf{Human: Intent Leakage ($\downarrow$)}} \\
\cmidrule(lr){1-4}
\textbf{Method} & ASR@1 & ASR@3 & MRR \\
\midrule
Soft-PO & 24.25 & 42.84 & 33.65 \\
GTKA   & 21.67 & 40.13 & 31.26 \\
\bottomrule
\end{tabular}
\caption{Human evaluation results on the BioQA dataset. $\uparrow$ indicates higher scores are better, while $\downarrow$ indicates lower scores are better.}
\label{tab:human_bioqa}
\end{table}

\subsection{Ablation study}
To comprehensively evaluate the contribution of each component in our proposed framework, we conducted ablation studies on the BioQA dataset using Qwen2.5-3B-Instruct as the local model and Qwen-Turbo as the external model. The results are reported in Table~\ref{tab:ablation_bioqa}, showing that each component contributes positively to the overall performance.

\subsection{Case Study}
We present a representative biomedical example to illustrate the advantage of our framework, as shown in Figure~\ref{fig:case_study}. Given the original query, directly querying an external LLM would expose the pathway name, key molecule, and specific cell type. Instead, our method generates several low-leakage sub-queries that deliberately stay at a general-mechanism level while still covering the essential reasoning steps from different angles. The external model answers these generic sub-queries without being exposed to the original intent, and a trusted local model then securely integrates them back into the original biological context to produce the final explanation. The integrated answer aligns closely with the ground truth, demonstrating that our framework can recover a highly consistent, context-specific response while minimizing intent leakage.

\begin{table}[htbp]
\centering
\resizebox{\columnwidth}{!}{
    \begin{tabular}{l|cc|cc}
    \toprule
    \textbf{Method} & \textbf{R-L} & \textbf{BertScore} & \textbf{ASR@3} & \textbf{MRR} \\
    \midrule
    GTKA (Full) & 16.42& 82.85& 36.14& 28.43\\ 
    \midrule
    \quad w/o DPO & 15.73& 82.19& 38.45& 30.12\\
    \quad w/o Quality ($\alpha=0$) & 14.47& 81.13& 33.24& 25.69\\
    \quad w/o Leakage ($\beta=0$)& 17.08& 83.07& 40.32& 33.41\\
    \bottomrule
    \end{tabular}
}
\caption{Ablation study on the BioQA dataset.}
\label{tab:ablation_bioqa}
\end{table}

\section{Conclusion}
In this work, we presented Game-theoretic Trustworthy Knowledge Acquisition (GTKA) to reconcile the need for the advanced and continually updated knowledge of external LLMs with privacy constraints. By formulating query decomposition as an adversarial game, our framework generates generalized sub-queries that maximize utility while minimizing leakage. Through extensive experiments on two newly constructed benchmarks in the biomedical and legal domains, we demonstrated that GTKA significantly outperforms existing baselines, achieving a superior balance of privacy and answer fidelity. Unlike heavy cryptographic solutions or infrastructure-level changes, GTKA offers a lightweight, semantic-level defense that allows users to safely access the evolving information in external models without exposing proprietary intent, while remaining compatible with today’s deployed LLM services and practical real-world workflows.

\section*{Limitations}
Our study presents two primary limitations:
\begin{itemize}
    \item While we have rigorously validated our method in the biomedical and legal domains, the framework's generalizability to other fields remains to be verified. Sectors such as finance, engineering, or open-domain scenarios possess distinct data distributions and privacy constraints. These differences could potentially shift the critical equilibrium between utility and leakage.
    \item  Our current approach dispatches sub-queries to a single external LLM provider. Theoretically, distributing sub-queries across multiple heterogeneous providers could offer stronger privacy guarantees. However, such a multi-provider strategy inevitably incurs higher computational overhead. We plan to explore optimization strategies that balance these resource demands with privacy gains in future work.
\end{itemize}

\section*{Ethical Statement}
All experiments in this study were conducted in a controlled, simulated laboratory environment. We explicitly state that the datasets utilized in our work are sourced exclusively from publicly available benchmarks and contain no Personally Identifiable Information. Furthermore, our adversarial assessments and privacy reconstruction attacks were strictly confined to local internal models. We did not perform any attacks against commercial external LLM servers or any third-party infrastructure.

For the human evaluation conducted, we recruited three domain experts. Prior to data collection, we explicitly explained the intended usage of the data to the annotators in detail. Regarding data consent, the BioQA and LawQA datasets constructed in this work utilize publicly available documents from PubMed and U.S. judicial decisions, respectively. These sources are open for research purposes.

\bibliography{custom}

\appendix

\section{Dataset Construction}
\label{app:dataset_construction}
For the BioQA dataset, we first collect 5{,}000 PubMed articles as source documents and use GPT-5 to generate three QA pairs per article, resulting in 15{,}000 raw QA pairs. To ensure data quality, we employ an LLM-based judge to score each QA pair on a 0--5 scale based on overall quality. We retain QA pairs with scores above 4 and further remove semantically redundant instances by filtering near-duplicate QA pairs. After these filtering steps, BioQA contains 12{,}876 QA pairs, with some illustrative examples shown in Table~\ref{tab:bioqa_examples}. 

For the LawQA dataset, we collect 15{,}000 U.S.\ judicial decisions and use GPT-5 to generate one QA pair per decision, producing 15{,}000 raw QA pairs. We apply the same quality-scoring and near-duplicate removal procedure, retaining only QA pairs with scores above 4 and removing highly similar instances, yielding 12{,}575 QA pairs in the final LawQA dataset. Some illustrative examples are provided in Table~\ref{tab:lawqa_examples}.

\begin{table*}[t]
\centering
\small
\setlength{\tabcolsep}{6pt}
\renewcommand{\arraystretch}{1.15}
\resizebox{\textwidth}{!}{
\begin{tabular}{p{0.05\textwidth} p{0.47\textwidth} p{0.48\textwidth}}
\toprule
ID & Question & Answer \\
\midrule
1 &
How do immune checkpoints like PD-1 and CTLA-4 suppress T-cell activation? &
PD-1 and CTLA-4 attenuate T-cell receptor signaling by recruiting phosphatases that dephosphorylate key signaling molecules, reducing PI3K/Akt activity, IL-2 production, and co-stimulatory signals, which collectively limit T-cell proliferation and effector function. \\
\midrule
2 &
How does NOD2 activation enhance dendritic cell maturation and antigen presentation in the context of vaccination? &
NOD2 signaling upregulates costimulatory molecules (such as CD80 and CD86), MHC class II, and cytokines that promote T cell priming. It also induces autophagy, which facilitates antigen processing and can support cross-presentation to CD8+ T cells. These changes increase the efficiency of antigen presentation and promote migration to lymphoid tissues via chemokine receptor induction. \\
\bottomrule
\end{tabular}
}
\caption{Illustrative examples from BioQA.}
\label{tab:bioqa_examples}
\end{table*}

\begin{table*}[t]
\centering
\small
\setlength{\tabcolsep}{6pt}
\renewcommand{\arraystretch}{1.15}
\resizebox{\textwidth}{!}{
\begin{tabular}{p{0.05\textwidth} p{0.47\textwidth} p{0.48\textwidth}}
\toprule
ID & Question & Answer \\
\midrule
1 &
In a Connecticut criminal matter where the court deferred the evidentiary portion of the defendant’s trial until the codefendant’s case concluded, I need to assess the viability of the defendant’s speedy-trial claim on appeal. Under Connecticut Practice Book § 3-10 and State v. Mooney, 218 Conn. 85 (1991), how should we frame the analysis given the record facts that the delay was caused by the prosecutor’s unforeseen back injury and that the defendant did not object to the deferral at trial? Please outline how these facts affect preservation and the merits, and what arguments the state can emphasize to show the delay was not sufficiently long and served the interest in a fair and impartial proceeding.  &
Applying Practice Book § 3-10 and Mooney’s balancing framework, the court weighed the defendant’s speedy-trial interest against the state’s interest in a fair, impartial trial and found no violation. Two facts are decisive. First, the delay was not significantly long, which undercuts the prejudice component of a speedy-trial claim. Second, the cause of delay—the prosecutor’s back injury—was unforeseen and beyond the state’s control, supporting the reasonableness of the deferral. On preservation, the defendant did not object to the delay at trial, so the speedy-trial claim was not preserved for appellate review. The state should emphasize that the deferral served the integrity of the proceedings, the length of delay was not substantial, the cause was legitimate and uncontrollable, and the absence of a contemporaneous objection bars the claim on appeal. \\
\midrule
2 &
On the limitations issue, where the alleged offenses occurred in 1949 and the indictment was not found until 1952, and a motion to dismiss has been raised on statute-of-limitations grounds under 18 U.S.C. § 3282 (three-year limit), confirm whether the indictment is time-barred and indicate the proper disposition (including whether dismissal should be without prejudice) based on how the court applied the timing rule here. &
Yes. Under 18 U.S.C. § 3282, an indictment must be found within three years after the commission of the offense. Because the offenses occurred in 1949 and the indictment was not found until 1952, it fell outside the three-year period and was time-barred. The court granted the motion to dismiss and dismissed the indictment without prejudice. \\
\bottomrule
\end{tabular}
}
\caption{Illustrative examples from LawQA.}
\label{tab:lawqa_examples}
\end{table*}

\section{Metric Calculations}
\label{app:metrics}
We use ASR@k and MRR to quantify intent leakage under an attacker who observes the privacy-preserved input and attempts to recover the true source segment from a candidate pool (the true segment plus decoys). ASR@k captures whether the attacker can successfully identify the true segment within the top-$k$ predictions, while MRR further reflects how highly the true segment is ranked across all candidates.

\paragraph{Attack Success Rate (ASR).}
For each test instance $i \in \{1,\dots,M\}$, let $\mathcal{C}_i=\{c_{i,1},\dots,c_{i,N}\}$ denote the candidate set containing exactly one true source segment $c_i^\star$ and $N-1$ decoy segments.
Let the attacker output a ranked list over $\mathcal{C}_i$, and let $r_i \in \{1,\dots,N\}$ denote the rank position of $c_i^\star$ (smaller is better).
We define ASR@k as:
\begin{equation}
\mathrm{ASR@}k = \frac{1}{M}\sum_{i=1}^{M} \mathbb{I}\left[r_i \le k\right],
\end{equation}
where $\mathbb{I}[\cdot]$ is the indicator function.

Intuitively, ASR@k measures the probability that the attacker places the true segment among its top-$k$ candidates. A higher ASR@k indicates stronger intent leakage.

\paragraph{Mean Reciprocal Rank (MRR).}
MRR reflects how highly the true source segment is ranked within the attacker’s ranked list over the candidate set $\mathcal{C}_i$.
A higher MRR indicates that the attacker tends to place the true segment closer to the top, suggesting stronger intent leakage, while a lower MRR indicates weaker leakage.
Formally, MRR is defined as:
\begin{equation}
\mathrm{MRR} = \frac{1}{M}\sum_{i=1}^{M} \frac{1}{r_i},
\end{equation}
where $r_i$ denotes the rank position of the true segment $c_i^\star$ for instance $i$, and MRR takes values in $(0,1]$.

\section{Experimental Settings}
\label{app:exp_setting}

Across all LLMs, we standardize the decoding hyperparameters with a sampling temperature of $0.7$, nucleus sampling ($\text{Top-}p$) of $0.9$, and a maximum generation length of $512$ tokens. Regarding the specific hyperparameters of GTKA, we set the number of candidate sub-query groups $K=4$ and the number of sub-queries per group $n=9$. The reward function balances quality and privacy with weights set to $\alpha=2/3$ and $\beta=1/3$, respectively, and the iterative adversarial training process is conducted for $5$ rounds. The generator is optimized via DPO with a learning rate of $5\text{e-}6$. The prompt templates utilized for generating sub-queries from the original query are detailed in Appendix~\ref{app:prompts}. For all comparative baselines, we strictly follow the optimal configurations reported in their original papers to ensure a fair comparison.

\section{Prompt Details}
\label{app:prompts}
The prompts used to generate sub-queries from the original queries on the BioQA and LawQA datasets are as follows:
\begin{figure}[htbp]
\setlength{\belowcaptionskip}{-0cm}
    \centering
    \includegraphics[width=1\linewidth]{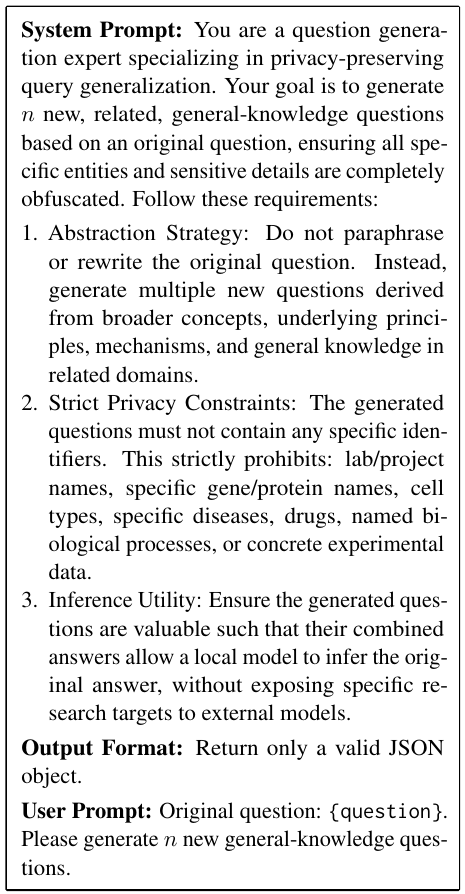}
    \caption{Prompt (BioQA).}
    \label{fig:prompt_bioqa}
\end{figure}

\begin{figure}[htbp]
\setlength{\belowcaptionskip}{-0cm}
    \centering
    \includegraphics[width=1\linewidth]{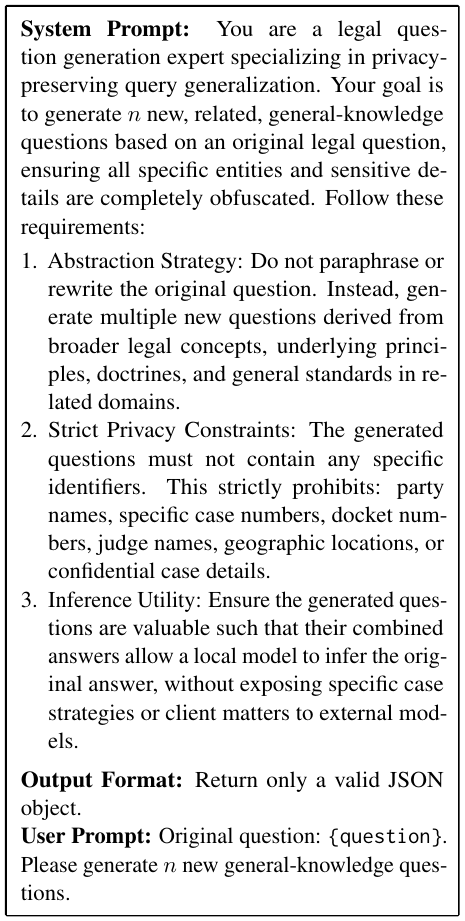}
    \caption{Prompt (LawQA).}
    \label{fig:prompt_lawqa}
\end{figure}

\end{document}